\documentclass[10pt,twocolumn,letterpaper]{article}

\usepackage[pagenumbers]{cvpr} 

\usepackage[percent]{overpic}
\usepackage{graphicx}
\usepackage{amsmath}
\usepackage{amssymb}
\usepackage{booktabs}
\usepackage{multirow}
\usepackage{afterpage}
\usepackage{subcaption}

\newcommand{\tabincell}[2]{\begin{tabular}{@{}#1@{}}#2\end{tabular}}  

\usepackage[pagebackref,breaklinks,colorlinks]{hyperref}

\usepackage[table,xcdraw]{xcolor}
\usepackage{fontawesome5}
\usepackage[capitalize]{cleveref}
\crefname{section}{Sec.}{Secs.}
\Crefname{section}{Section}{Sections}
\Crefname{table}{Table}{Tables}
\crefname{table}{Tab.}{Tabs.}

\usepackage{xcolor}

\definecolor{electricindigo}{rgb}{0.44, 0.0, 1.0}
\definecolor{lightblue}{RGB}{240,245,255}
\definecolor{darkblue}{RGB}{40,40,85}
\definecolor{babyblue}{rgb}{0.54, 0.81, 0.94}
\definecolor{pearDark}{HTML}{2980B9}
\definecolor{pearDarker}{HTML}{1D2DEC}

\def\cvprPaperID{8233} 
\def\confName{CVPR}
\def\confYear{2023}

\begin{document}

\title{TrafficCAM: A Versatile Dataset for Traffic Flow Segmentation}

\author{Zhongying Deng\textsuperscript{1,2}
,
Yanqi Chen\textsuperscript{2}
, 
Lihao Liu\textsuperscript{2}
, 
Shujun Wang\textsuperscript{2}
, 
Rihuan Ke\textsuperscript{3}
, \\
Carola-Bibiane Schönlieb\textsuperscript{2}
, 
Angelica I Aviles-Rivero\textsuperscript{2} \\
\textsuperscript{1}University of Surrey, UK $\quad$
\textsuperscript{2}University of Cambridge, UK $\quad$
\textsuperscript{3}University of Bristol, UK\\
}



\twocolumn[{%
    \renewcommand\twocolumn[1][]{#1}%
    \maketitle
    \begin{center}
      \vspace*{-15pt}
      \begin{overpic}[width=\textwidth]{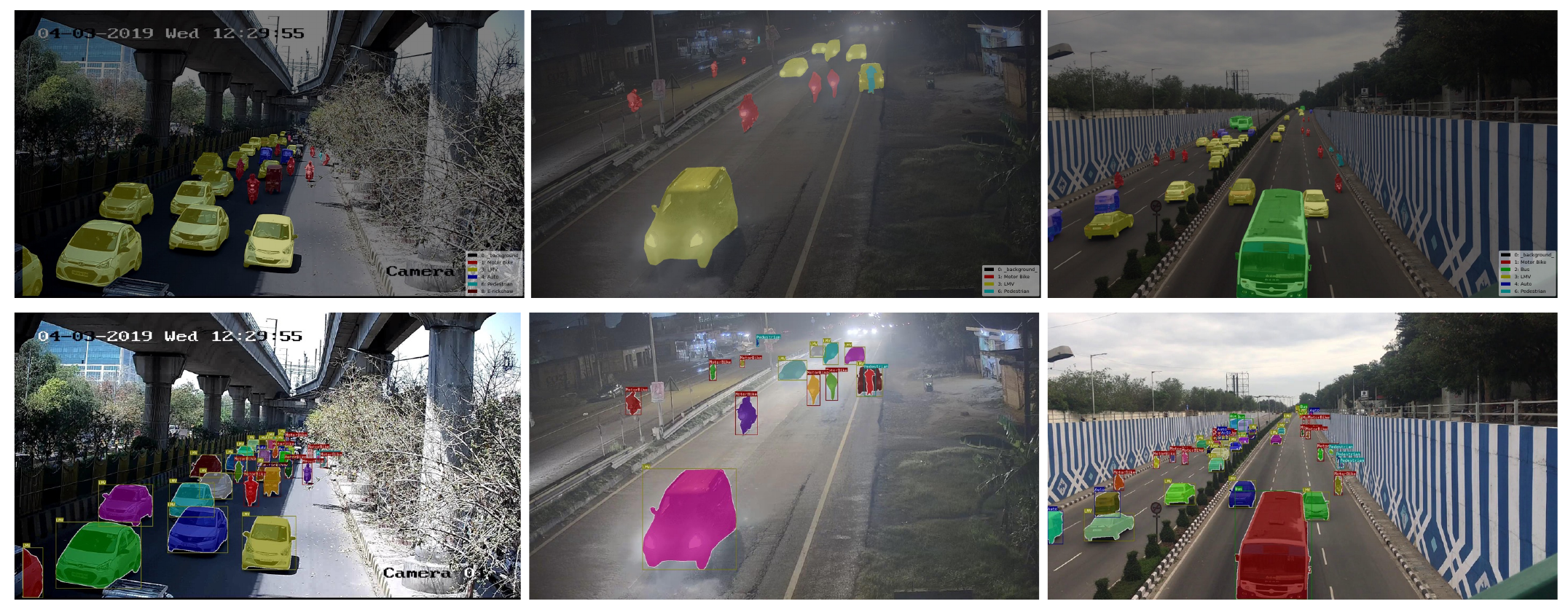}
      \end{overpic}
      
      \vspace*{-5pt}
    \end{center}%
}]


\begin{abstract}

Traffic flow analysis is revolutionising traffic management.
Qualifying traffic flow data, traffic control bureaus could provide drivers with real-time alerts, advising the fastest routes and therefore optimising transportation logistics and reducing congestion. The existing traffic flow datasets have two major limitations. They feature a limited number of classes, usually limited to one type of vehicle, and the scarcity of unlabelled data.
In this paper, we introduce  a new benchmark traffic flow image dataset called TrafficCAM. Our dataset  distinguishes itself by two major highlights. Firstly, TrafficCAM provides both  pixel-level and instance-level semantic labelling along with  a large range of types of vehicles and pedestrians. 
It is composed of a large and diverse set of video sequences recorded in streets from eight Indian cities with stationary cameras.
Secondly, TrafficCAM aims to establish a new benchmark for developing fully-supervised tasks, and importantly, semi-supervised learning techniques. It is the first dataset that provides a vast amount of unlabelled data, helping to better capture traffic flow qualification under a low cost annotation requirement.
%
More precisely, our dataset has 4,402 image frames with semantic and instance annotations along with 59,944 unlabelled image frames.
%
%
We validate our new dataset through a large and comprehensive range of experiments on
several state-of-the-art approaches under four different settings: fully-supervised semantic and instance segmentation, and semi-supervised semantic and instance segmentation tasks. 
%
Our benchmark dataset will be released in \url{https://anonymous_submission}.

\end{abstract}

\section{Introduction}
\label{sec:intro}







With the rapid  population growth, the increasing number of vehicles~\cite{downs1962law}, 
and 
human repetitive travel patterns~\cite{gonzalez2008understanding},  traffic congestion in city networks has become challenging to analyse~\cite{loder2019understanding}.
%
The overcrowding of road traffic increases the risk of traffic accidents 
and air pollution which undermines environmental sustainability.
For that reason, smart traffic solutions are a great focus of interest. 
Using smart traffic solutions, traffic control bureau could provide drivers with real-time alerts advising the
fastest routes, and therefore optimising transportation logistics and reducing congestion.
However, deep learning-based research to develop those smart solutions requires extensive traffic data analysis 
to find patterns. 
%


There is a large number of datasets for city/street scene understanding with moving and low position cameras including Cityscapes \cite{cordts2016cityscapes}, Mapillary \cite{neuhold2017mapillary} and IDD \cite{varma2019idd}. However, their purpose greatly differs from ours. Contrary to these datasets that seek to provide understanding of urban level street, with classes such as buildings, vegetation and street objects, our TrafficCAM dataset seeks to focus on the traffic flow analysis. That is, our setting is data coming from  traffic cameras instead of a moving street level camera. Moreover, our dataset provides more classes for different types of vehicles.

Machine learning techniques have proven to be very successful for pattern recognition~\cite{wang2019deep} based on the requirement of large amounts of data.
In the field of traffic data, this however is a limiting factor. Currently available datasets are either very small (see e.g., \cite{wang2008unsupervised, tang2019cityflow}) or make use of moving cameras (see e.g., \cite{cordts2016cityscapes, xia2018dota}), which makes traffic flow analysis nearly impossible. \textit{Traffic flow analysis from images and videos boils down to segmenting vehicles and people from surroundings.}
We thus introduce the new, large, fixed camera traffic dataset named TrafficCAM. TrafficCAM not only covers various traffic scenes but also contains sufficient challenging samples to provide a solid basis for traffic flow segmentation (see teaser).  Our contributions are:

%
\faHandPointRight[regular] We introduce the largest benchmark for traffic 
segmentation covering a wide range of cities in India.
TrafficCAM dataset is specifically designed for traffic-related object segmentation with both pixel and instance annotations. Unlike existing datasets for flow analysis, we provide the first dataset with a large variety of vehicles.

\smallskip
\faHandPointRight[regular] Our TrafficCAM dataset, 
and different from other datasets, is composed of a vast amount of unlabelled frames.
This key feature opens the door to further developments of robust and generalisable models that 
learn with limited annotated data and a large number of unlabelled data.

\smallskip
\faHandPointRight[regular] We validate the usability of our dataset through a set of extensive experiments over the existing
state-of-the-art methods, and for four different settings supervised learning to semi-supervised learning for both semantic and instance segmentation.


%
%
%
%
%
%
\section{Related Work} \label{sec:relatedwork}
Traffic data has been an inherent part of research in traffic management since nearly twenty years. In general, datasets in this area can be classified into two main categories -- fixed camera traffic data and moving camera traffic data. Whilst the latter is captured e.g., from within moving cars, fixed camera traffic datasets make use of static cameras recording scenes within the same surroundings. 


Moving camera datasets are popular in the community, where the goal is to  record scenes with a non-static camera which can be attached to a car. 
Within this category several datasets have been introduced. 
The probably most famous dataset in this category is Cityscapes~\cite{cordts2016cityscapes} which has been introduced in 2016. Further datasets in this area include the Honda Research Institute Driving Dataset (HDD)~\cite{ramanishka2018toward}, the Dataset for Object deTection in Aerial images (DOTA)~\cite{xia2018dota} and the IDD dataset~\cite{varma2019idd}. We underline that this family of datasets greatly differs from ours and our purpose. Their goal is to capture not only traffic flow but general urban level objects. For example, buildings, vegetation, animals and non-traffic vehicles such as boats.  Whilst our purpose is tackling inherent factors for traffic flow including different type of vehicles and pedestrians.
%
%
Our TrafficCAM is closer in purpose to existing fixed camera traffic datasets.  In the following, we review the existing traffic datasets following this philosophy.

\subsection{Traffic Datasets with Fixed Camera }
Fixed camera traffic datasets use stationary cameras to film scenes within the same surroundings. Most of the time, the cameras are installed several meters above street level, e.g., next to traffic lights. 
These types of datasets can be broadly classified into categories of single-  and multi- devices datasets. Single device datasets make use of a single camera, whilst multi device datasets use either multiple cameras along with other sensors like laser scanners or thermal infrared cameras. 


For traffic flow analysis, there exist two fixed single camera datasets. To the best of our knowledge, the oldest available dataset is the Highway Traffic Videos~\cite{jodoin2014urban} dataset. It was published in 2004 and consists of two days of video footage of a CCTV camera in Seattle, USA. Another dataset  is the MIT Traffic~\cite{wang2008unsupervised} which has been introduced in 2009 provides a total of 90 minutes of video clips, and it is splitted into 20 sequences.

More recent datasets make use of multiple devices, either by recording videos of multiple cameras or a camera along with another sensory device. Multi camera traffic datasets include Urban Tracker~\cite{jodoin2014urban} and CityFlow~\cite{tang2019cityflow}. Urban Tracker uses five cameras to film different surroundings from different angles. It has been introduced in 2014 and not only provides videos for traffic flow analysis of motorised vehicles but also includes pedestrians.
For the CityFlow dataset, three hours of video of ten intersections were filmed by 40 cameras. It was published in 2019.

In addition to only multiple cameras, the Ko-PER dataset~\cite{strigel2014ko} and AAU RainSnow Traffic Surveillance dataset~\cite{AAURainSnow} incorporate additional sensory data. Ko-PER uses eight monochromal cameras and 14 laser scanners to film a single intersection. The AAU RainSnow dataset however uses seven RGB cameras and seven thermal infrared cameras to film 22 five minute videos of seven intersections in Denmark.

 \begin{table*} [!h]
    \centering
    \caption{Comparison of our dataset and existing traffic surveillance datasets. Video duration column shows the number of videos and the total video duration. \ddag: released in 2023 but acquired in 2019.  
    }
    \label{tab:dataset_comparison}
        \resizebox{1\textwidth}{!}
    {
        \setlength\tabcolsep{1.2pt}
        \centering
        \begin{tabular}{l|cccccccc}
            \toprule[1pt]
            \cellcolor[HTML]{EFEFEF}{\textsc{Dataset}} & \cellcolor[HTML]{EFEFEF}{Year}& \cellcolor[HTML]{EFEFEF}{Videos Duration}  & \cellcolor[HTML]{EFEFEF}{\#Frames}  & \cellcolor[HTML]{EFEFEF}{\#Classes } & \cellcolor[HTML]{EFEFEF}{Resolution} &  \cellcolor[HTML]{EFEFEF}{Annotation Type} & \cellcolor[HTML]{EFEFEF}Task & \cellcolor[HTML]{EFEFEF}\tabincell{c}{Extra Frames\\ w/o Annotations}\\
            \toprule[1pt]
            MIT Traffic~\cite{wang2008unsupervised} & 2008 & 1/90 mins & - & 1 & 720$\times$ 480 & Bounding box & Detection &-\\
                        \hline
            UrbanTracker \cite{jodoin2014urban}& 2014 &5/- & 8,141& 3& \tabincell{c}{800$\times$600\\ 1024$\times$576\\1280$\times$720} & Bounding box & Tracking &-\\
                        \hline
                        
            Ko-PER~\cite{strigel2014ko} &2014&6.28 mins&-&1&656$\times$494 &Bounding box & Tracking &-\\
            \hline
            
            AAU RainSnow \cite{AAURainSnow}&2018&22/109 mins&2,200&3&640$\times$480&Pixel level& \tabincell{c}{Semantic \\Segmentation}&-\\
            \hline
            CityFlow~\cite{tang2019cityflow} & 2019 & 40/195 mins & - &1 &  960p& Bounding box & Tracking &-\\
                        \hline
            
            TrafficCAM (Ours)& 2023\textsuperscript{\ddag} & - & 4,402 & 10 &
            \tabincell{c}{1920$\times$1080\\ 1289$\times$720\\1056$\times$864 \\352$\times$288} & \tabincell{c}{Pixel level\\Instance level} & \tabincell{c}{Semantic \\ \& Instance \\Segmentation} &59,944\\

            
            \toprule[1pt]
        \end{tabular}
    }
\end{table*}

\subsection{Existing Datasets \& Comparison to Ours}
TrafficCAM is a unique dataset within its category. In contrast to existing ones~\cite{wang2008unsupervised,jodoin2014urban,strigel2014ko,AAURainSnow,tang2019cityflow}, TrafficCAM provides 10 classes whilst the majority of datasets solely focus on a single class.  This feature provides an opportunity to get a more detailed \textit{uninterrupted flow} analysis~\cite{bae2019spatio,li2017vehicle}, where more insightful vehicle-vehicle types analysis can be achieved. This is translated to have more complex yet informative models for better traffic policies.

Secondly, TrafficCAM is the first dataset to highlight a large amount of unlabelled data for the model development. Existing datasets are limited to feature annotated data, however, labelling is expensive and time consuming. By introducing a vast amount of unlabelled data we open the door to more robust and generalisable models that lean with limited annotations. A summary of our key dataset highlights are summarised in Table~\ref{tab:dataset_comparison}.

\section{TrafficCAM: A New Dataset for Segmentation}
\label{sec:dataset}


The TrafficCAM dataset is collected to 
capture the traffic flow from complex city networks  in  India. 
Although designing such a large-scale dataset requires a lot of efforts on image collection and annotations, our dataset has the potential to promote the traffic solution with respect to  smart cities.
%
%
%
%
In what follows, we detail the characteristics of  
our TrafficCAM dataset according to the data specifications, classes and annotations, and statistical analysis.

\subsection{Data Specifications}

\begin{figure}[t!]
\centering
\begin{subfigure}{.5\columnwidth}
  \centering
  \includegraphics[width=.9\columnwidth]{ 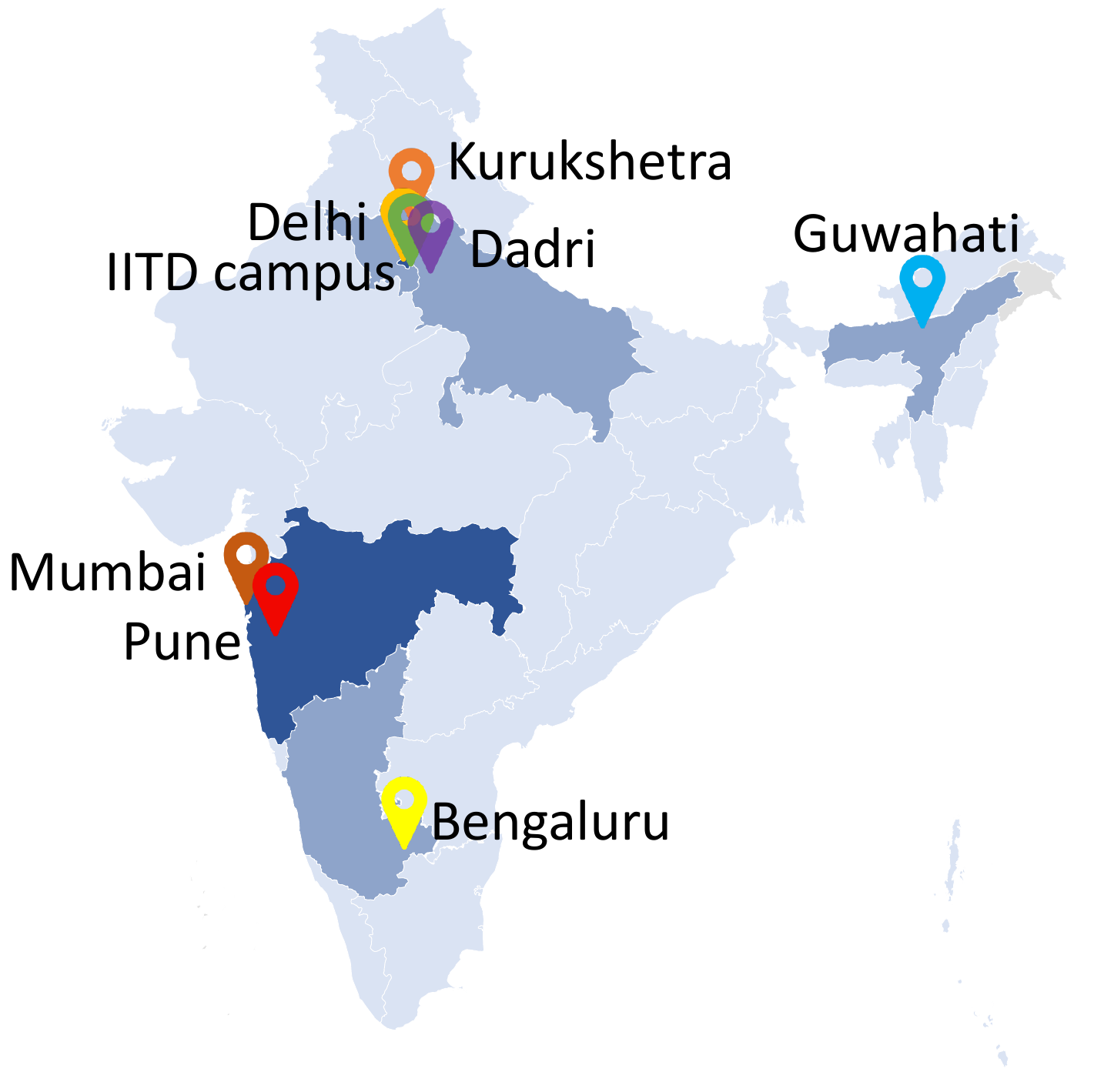}
  \caption{}
  \label{fig:plot_map}
\end{subfigure}%
\begin{subfigure}{.55\columnwidth}
  \centering
  \includegraphics[width=.9\linewidth]{ 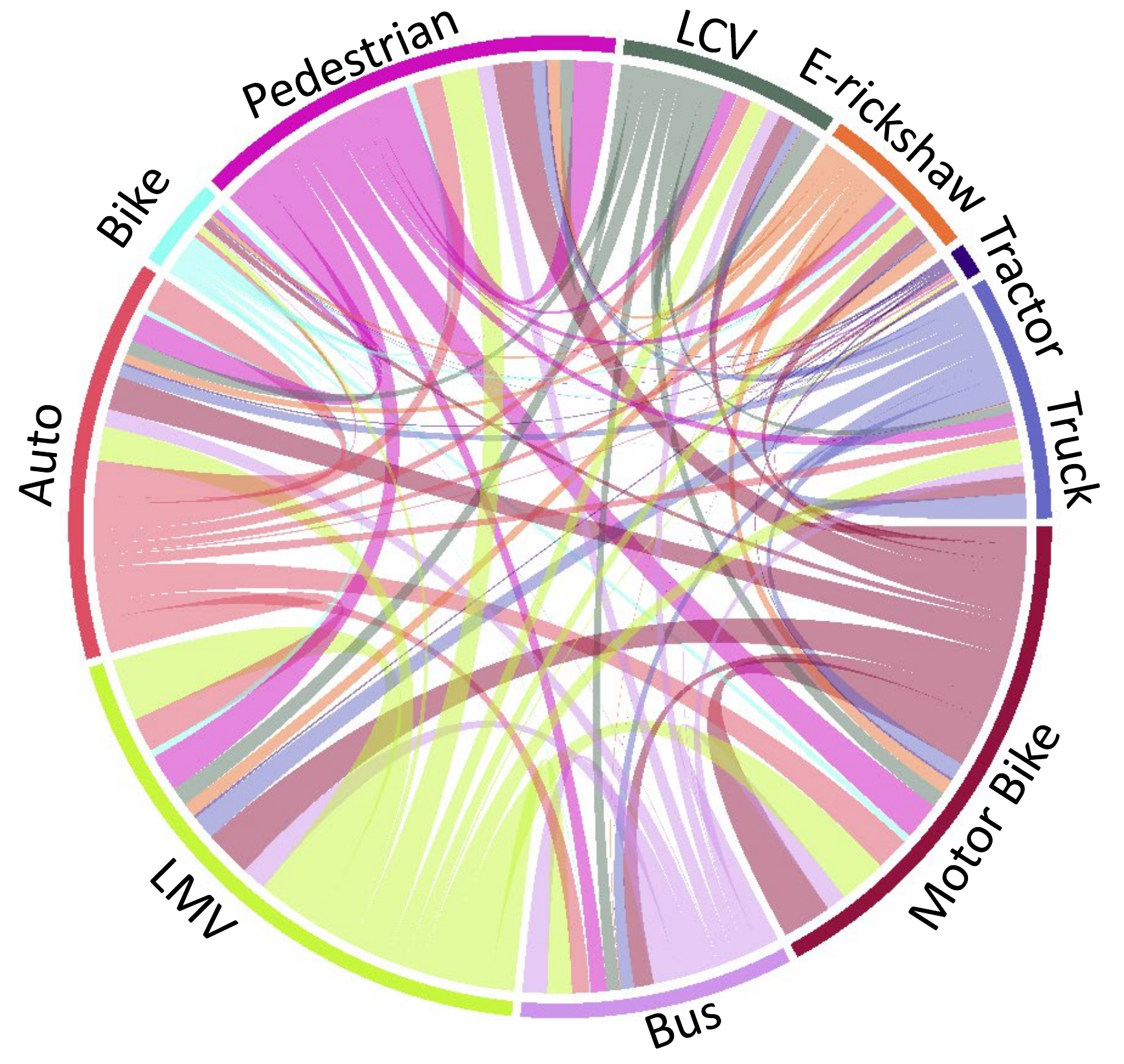}
  \caption{}
  \label{fig:plot_correlation}
\end{subfigure}
\caption{(a) Geographic distribution of TrafficCAM collection. (b) Correlation analysis.}
\end{figure}

The  TrafficCAM dataset is collected from the traffic stationary cameras from eight cities in India.
We detail the city locations  in Figure~\ref{fig:plot_map}.
This is a good indication that the image frames in the TrafficCAM dataset are from diverse regions, including the geographic north, south, inner, west, and east parts, which is a distinctive feature of the proposed benchmark.
The videos in our TrafficCAM dataset are captured with various types of cameras, so that our data is not limited to one resolution.
The resolution of our collected data ranges from $352\times288$ to $1920 \times 1080$. More information can be found in Table~\ref{tab:dataset_comparison}.

%
%
%
The TrafficCAM dataset is composed of 2,148 videos, where we extract a total of  64,346 frames.
Specifically, the frames from 78 videos have fully pixel-level annotations (76 have 30 frames per video, 1 has 29 frames and 1 has 25 frames), while 2,068 videos only have the first frame annotation. 
%
Therefore, our TrafficCAM dataset has a large number of frames without annotations (59944 frames) from the above partially annotated videos, which opens the door to the development of techniques relying on limited annotations as semi-supervised learning.
%


We split our TrafficCAM dataset into  training, validation, and test sets. 
The dataset is split at the video level, which guarantees that the frames from the same video  only appear in one split. 
We chose not to split the data randomly due to the special annotation character of our TrafficCAM dataset.
%
Specifically, we set all the videos with only first-frame annotation and 8 fully annotated videos (30 frames per video) as the training split, while the remaining videos with fully annotations are  randomly split into validation and testing. We left-out 12 videos for future purposes like optical flow learning.
Our split criteria finally leads to 2,308 annotated frames, and 59,944 unannotated frames for training, 210 annotated frames for validation, and 1,524 annotated frames for testing.
We also remark that our dataset covers several conditions such as night and day acquisition, and other conditions such as foggy scenarios.

\subsection{Classes and Annotations Details}
\begin{figure*}[!t]
\centering
\includegraphics[width=\textwidth]{ 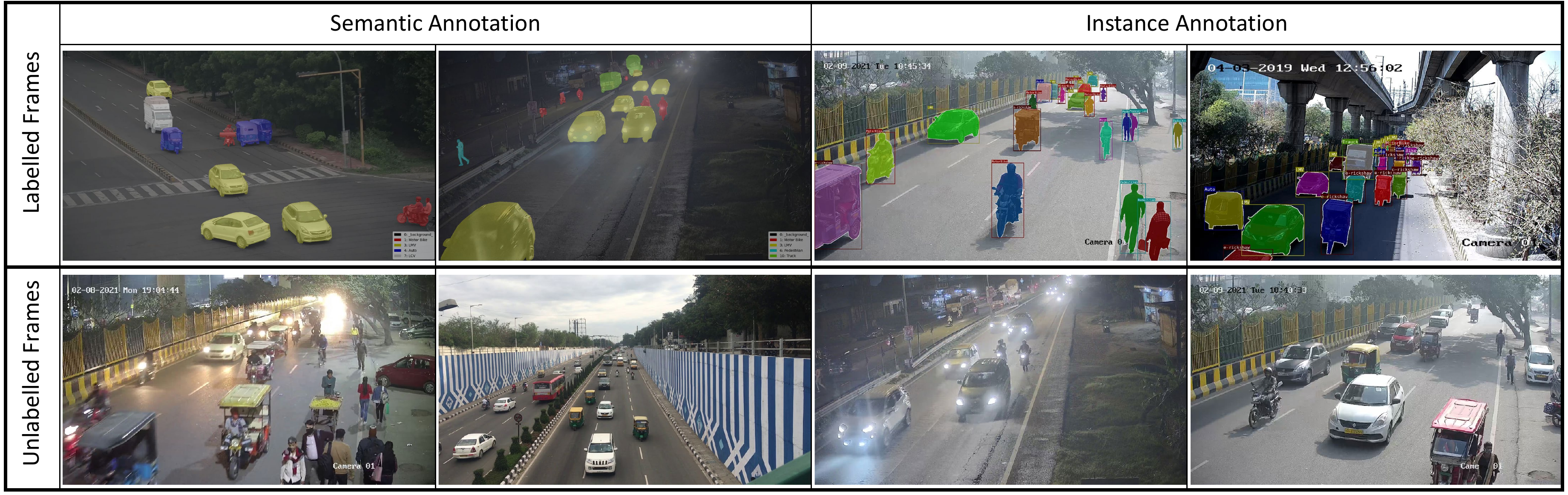} 
\caption{Illustration of our TrafficCAM dataset. The first row shows labelled frames, where the first two columns are with semantic annotation and latter two with instance annotation frames. The second row consists of unlabelled image frames.}
\label{fig:more_example}
\end{figure*}

For each annotated frame in TrafficCAM dataset, we provide pixel annotations for ten classes in both semantic and instance levels.
Due to the attribute of our data to analyse traffic flow data, \textit{we only provide annotations on the moving traffic vehicles and pedestrians on the main road. }
More annotation examples and unannotated data can be found in Figure~\ref{fig:more_example}.

\begin{figure}[!t]
\centering
\includegraphics[width=0.45\textwidth]{ 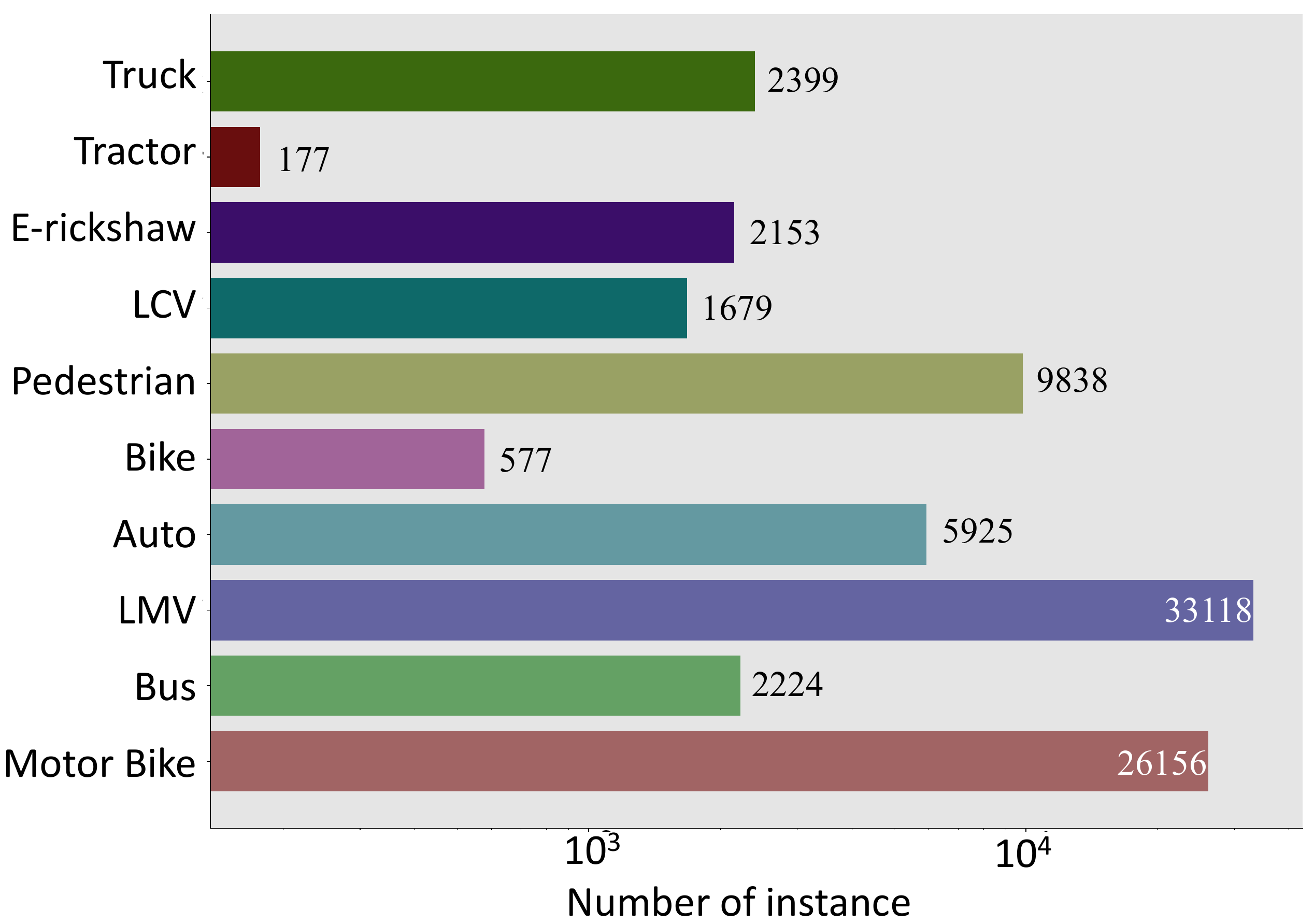} 
\caption{Illustration of number of images per category.}
\label{fig:plot_class}
\end{figure}

The dataset 
primarily affects the image content and the traffic categories. 
For our collected TrafficCAM dataset, we distinguish among 10 categories, which can be organised into 2 groups, vehicle and pedestrian. 
The vehicle group contains 9 different classes, namely Motor Bike, Bus, LMV, Auto, Bike, LCV, E-rickshaw, Tractor, and Truck.
The class overview is shown in Figure~\ref{fig:plot_class}. We highlight that this is the first traffic flow dataset to provide such large variety of classes.

The criteria for selecting categories for the annotations process are driven by the following reasons.
1) Our dataset mainly focuses on the traffic information instead of scene understanding like Cityscapes \cite{cordts2016cityscapes}. Therefore, vehicles and people play an important role.
2) We  provide a large variety of vehicles captured from a large range of cities in India, where unique and complex vehicles and traffic situations can be captured. This feature introduces a unique opportunity to develop more robust techniques.
The most common types of cars on Indian streets have been included under this category.

\textbf{Frames without Annotations.}
Another key highlight in our dataset is that we provide a massive amount of unlabelled samples making
our TrafficCAM dataset substantially different from other existing datasets.
The main purpose to provide such a large unlabelled set is two-fold. Firstly, annotations are time-consuming and expensive.
Secondly, our TrafficCAM dataset provides an opportunity for developing robust and generalisable techniques  with limited annotated data and a large number of unlabelled data.
We therefore provide a strong semi-supervised benchmark with the ratio of unlabelled frames to labelled frames reaching 30:1.
%
In the experiment section, we provide comparison results not only under the fully supervised learning setting but also under semi-supervised learning schemes.
\textit{To the best of our knowledge, TrafficCAM is the largest traffic object segmentation benchmark under both supervised and semi-supervised settings.}

\subsection{Statistical Analysis}

We illustrate the number of images per category in Figure~\ref{fig:plot_class}.
As we can see, our dataset is a typical long-tailed dataset. 
The Motor Bike and LMV occupy around $90\%$ percentage of all frames, while the frame numbers of Bike and Tractor are no more than $600$.
Such imbalance natural character makes our TrafficCAM dataset more challenging yet of a great interest for model development.

\begin{figure}[!t]
\centering
\includegraphics[width=0.48\textwidth]{ 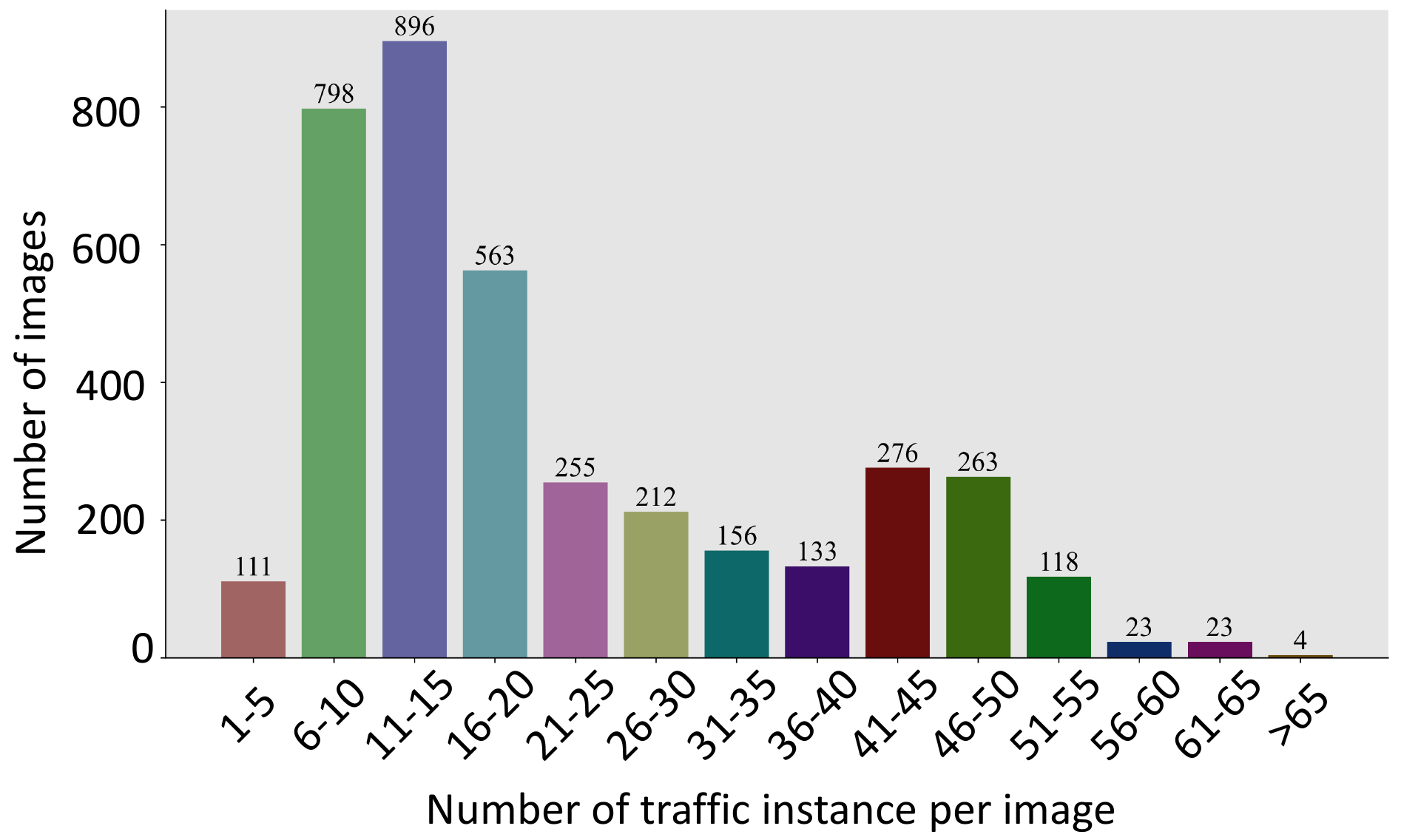} 
\caption{Dataset statistics regarding complexity: the frequency of images with a fixed number of annotated object instances per image.}
\label{fig:plot_instance}
\end{figure}

We report in Figure~\ref{fig:plot_instance} the frequency of images with a fixed number of annotated objects in the TrafficCAM dataset.
TrafficCAM is the largest dataset focused on traffic stationary views. 
We find that our dataset covers a good variety of category complexity.
There are more than 600 images with the number of instances being more than 40 yielding to challenging cases. 
From Figure~\ref{fig:plot_correlation} we may also observe the diversity of classes in our dataset frames.

Compared with other existing traffic surveillance dataset, our TrafficCAM dataset contains the most video and frame numbers, as shown in Table~\ref{tab:dataset_comparison}.
The resolution of our dataset ranges in a big difference, which shows the diversity of the TrafficCAM dataset.
\textit{Furthermore, our dataset is the only one that provides both pixel and instance level annotations under traffic object segmentation scenario.}




\newcommand{\dataset}{TrafficCAM}
\section{Experimental Settings}

In this section, we validate our TrafficCAM dataset through a large and comprehensive range of experiments on
several state-of-the-art approaches under four different settings: fully-supervised semantic and instance segmentation, and semi-supervised semantic and instance segmentation tasks. 

Specifically, we highlight the semi-supervised tasks which aim to improve the model's performance using limited labelled data and enormous unlabelled set. This task is meaningful because it can significantly reduce the annotation cost -- annotating images is labor-extensive and time-consuming, especially when pixel-level annotations are required for segmentation tasks. 
Our dataset accommodates with this scenario as it contains both labelled and massive unlabelled data. 
Therefore, we further investigate different baseline methods on our dataset for semi-supervised semantic and instance segmentation.

\subsection{Evaluation Metrics}
\textbf{Semantic segmentation task: }
We use Intersection over Union scores (IoU) to evaluate semantic segmentation tasks on our benchmark. 
IoU is defined as $\frac{TP}{TP+FP+FN}$, where $TP$ stands for the number of true positive pixels, $FP$ for false positive, and $FN$ for false negative. 
We compute the IoU for each class and then report the mean IoU (mIoU) as the final results.

\textbf{Instance segmentation task: }
The evaluation metrics used for instance segmentation is region-level average precision (AP) for each class, the same as ~\cite{lin2014microsoft}. Specifically, the AP is calculated by first computing the IoU of a single instance in a region, then applying different thresholds, \eg, from 0.5 to 0.95 with step 0.05, to the IoU to obtain the scores for predictions, and finally averaging these scores. The class-wise AP is then used for computing mAP, which is the mean of the class-wise AP over all the classes. We also report the mAP$^{50\%}$ for the threshold=50\%.


\subsection{Fully Supervised Semantic Segmentation}
\label{sec:semantic_seg}
For this task defined on \dataset{}, we perform pixel-level classification for semantic labels, without considering multiple instances. For example, if multiple pedestrians are  in the same image, their pixels will be classified into the same label `pedestrian'.

\textbf{Experimental settings.}
We adopt MMSegmentation~\cite{mmseg2020} to run the experiments on our \dataset{}. During training, we adopt random flipping and scale as the data augmentation, with a scale ratio in $[0.5,2]$. We then crop the image to the same size for model training. 
We train the model using SGD optimiser with learning rate initially set as $0.01$ with polynomial decay at power of 0.9; momentum at $0.9$ and weight decay at $0.0005$.
All the baselines are trained with batch size of $16$ for $20K$ iterations, and tested using the model with the best validation mIoU result.

During inference, we flip and resize the testing images to multiple scales, the same as training. These multi-scale predictions are interpolated in a bi-linear way so that they are of the same size as the input images. The predictions of different scales are then averaged as the final prediction.

\textbf{State-of-the-art methods. }
We consider two different categories of most widely-used baselines: ResNet-based and Transformer-based methods. The most classical methods are usually based on ResNet, including FCN~\cite{long2015fully}, PSPNet~\cite{zhao2017pyramid}, DeepLabV3+~\cite{chen2018encoder}. FCN is the first work to exploit deep CNNs for semantic segmentation. It motivates the follow-up methods such as PSPNet and DeepLabV3+. The latter two methods try to improve FCN by constructing multi-scale representations. 
The latest methods are usually based on transformer. Transformers have achieved state-of-the-art performance on semantic segmentation task partly because the attention mechanism can effectively capture long-range dependencies for such task. Some representative works are SegFormer~\cite{xie2021segformer} and SETR~\cite{zheng2021rethinking}, both employing vision transformers (ViT)~\cite{dosovitskiy2020image} as backbone.

\textbf{Experimental results. }
We conduct experiments with different backbones and crop sizes to better evaluate the SOTA methods on \dataset{}. We show the results in Table~\ref{tab:semantic_fully} and find that DeepLabV3+ with ResNet-101 backbone and 512$\times$1024 crop size achieves the best performance, \ie, 66.49\%. 
In addition, we have following observations:

\faHandPointRight[regular] For these ResNet-based methods, deeper ResNet always achieves better mIoU regardless of different methods, \eg, FCN with ResNet-101 increases the mIoU of ResNet-50 by 7.85\% while DeepLabV3+ benefits from a deeper backbone by about 10\%. 
Crop sizes of 769$\times$769 usually brings a performance gain, compared with 512$\times$1024. For instance, it improves the performance of PSPNet from 63.61 to 66.21. The exception is that 769$\times$769 results in a subtle decrease for DeepLabV3+. 

\faHandPointRight[regular] For the transformer-based methods, we adopt a fixed crop size with 1:1 aspect ratio to facilitate patch embedding. For SegFormer, it proposes MiT backbone which is based on ViT but tailored for semantic segmentation tasks. It is clear that larger backbone models usually contribute to better mIoU, \eg, 64.01\% of MiT-B5 over 60.93\% of MiT-B4. SETR adopts ViT-Large as encoder and further designs different decoders, namely, the naive decoder (Naive), the progressive upsampling one (PUP), and the multi-level feature aggregation (MLA). We find that the naive version achieves the best performance on \dataset{}.

\faHandPointRight[regular] Comparing the ResNet-based methods with the transformer-based ones, we observe that the ResNet-101-based FCN, PSPNet, and DeepLabV3+ generally obtain better (or at least comparable) performance than the transformers. Notably, the worst one based on ResNet-101 is FCN with crop size of 512$\times$1024, achieving 63.46\%. This result is comparable with the best one based on transformer, \ie, SegFormer with MiT-B5 as backbone.

Overall, all these methods achieve unsatisfactory on our \dataset{}, demonstrating that \dataset{} is a challenging dataset for semantic segmentation. We owe the challenge to the diverse scenes and highly imbalanced class distributions.

\begin{table}
\centering
 \caption{Results for fully supervised semantic segmentation tasks. The best result is in bold.
 }
\resizebox{0.95\columnwidth}{!}{
\begin{tabular}{l|l|l|l|l }
\toprule
\textbf{}    & \cellcolor[HTML]{EFEFEF}\textbf{Methods} & \cellcolor[HTML]{EFEFEF}\textbf{Backbone} & \cellcolor[HTML]{EFEFEF}\textbf{Crop size} & \cellcolor[HTML]{EFEFEF}\textbf{mIoU}  \\
\hline
\multirow{9}{*}{\rotatebox{90}{ResNet}}       & \multirow{3}{*}{FCN~\cite{long2015fully}}            & R50              & $512 \times 1024$   & 55.61          \\
             &              & R101             & $512 \times 1024$   & 63.46          \\
             &              & R101             & $769 \times 769$    & 65.77          \\
            \cline{2-5}
            & \multirow{3}{*}{PSPNet~\cite{zhao2017pyramid}}          & R50              & $512 \times 1024$   & 60.58          \\
             &              & R101             & $512 \times 1024$   & 63.61          \\
             &              & R101             & $769 \times 769$    & 66.21          \\
             \cline{2-5}
             & \multirow{3}{*}{DeepLabV3+~\cite{chen2018encoder}}      & R50              & $512 \times 1024$   & 56.50           \\
             &              & R101             & $512 \times 1024$   & \textbf{66.49} \\
             &              & R101             & $769\times 769$     & 66.26          \\
             
\hline
\multirow{6}{*}{\rotatebox{90}{Transformers}} & \multirow{3}{*}{SegFormer~\cite{xie2021segformer}}       & MiT-B0           & $512 \times 512$    & 52.61          \\
             &              & MiT-B4           & $512 \times 512$    & 60.93          \\
             &              & MiT-B5           & $512 \times 512$    & 64.01          \\
             \cline{2-5}
             & \multirow{3}{*}{SETR~\cite{zheng2021rethinking}}            & ViT-L\_MLA       & $512 \times 512$    & 54.57          \\
             &              & ViT-L\_Naive     & $512 \times 512$    & 57.91          \\
             &              & ViT-L\_PUP       & $512 \times 512$    & 57.15 \\
\bottomrule
\end{tabular}
}
 \label{tab:semantic_fully}
 \\[-0.51cm]
\end{table}

\subsection{Fully Supervised Instance Segmentation}
Instance segmentation aims to detect and segment each object instance. In this task, even if these instance are from the same class, we will need to assign them a separate label.

\textbf{Experimental settings. }
\label{sec:setting_inst_fully}
The fully supervised instance segmentation baselines are implemented in MMDetection~\cite{mmdetection}. We adopt similar settings as in \cref{sec:semantic_seg}. The differences are that 1) we keep the crop size to $1333 \times 800$ for all the different instance segmentation methods; 2) we train the model using AdamW~\cite{loshchilov2017decoupled} optimizer with learning rate initially set as $0.0002$, weight decay at $0.0005$, and polynomial decay at power of $0.9$. All the methods are trained for $20K$ iterations.



\textbf{State-of-the-art methods. }
According to whether region proposals are used, the SOTA methods can be divided into one-stage (without) and two-stage (with region proposals) methods. 
We evaluate six SOTA methods on our \dataset{}, covering both one-stage and two-stage methods. They are:  
Mask R-CNN~\cite{He_2017}, %
Cascaded Mask R-CNN~\cite{Cai_2019}, %
InstaBoost~\cite{fang2019instaboost}, %
SOLOv2~\cite{wang2020solov2}, %
QueryInst~\cite{Fang_2021_ICCV}, %
and Mask2Former~\cite{cheng2021mask2former}.%

Mask R-CNN is a well-known instance segmentation method. It essentially exploits Faster R-CNN~\cite{ren2015faster} for instance detection and then uses FCN to segment each detected instance. 
Cascaded Mask R-CNN further improves it by using a sequence of detectors to balance the positive and negative samples. These two are two-stage methods which use region proposals for instance segmentation. 
InstaBoost uses the existing instance mask annotations for random jittering to objects so that the training set can be augmented. 
SOLOV2 introduces dynamic parameters to the mask head of object segmenter so that the mask head is conditioned on the location, where the location information usually contributes to better performance. Notably, SOLOv2 is a one-stage method without region proposal. 
QueryInst also employs a dynamic design, but differs in treating instances of interest as learnable queries. 
Mask2Former constrains cross-attention within predicted mask regions to extract localized features. It takes Swin transformer~\cite{liu2021swin} as its backbone.

\textbf{Experimental results. }
From Table~\ref{tab:instance_fully}, we can observe that 1) Mask2Former equipped with Swin-S~\cite{liu2021swin} and InstaBoost with ResNet-101 outperform all the other competitors, both ranking first in terms of mAP. Mask2Former also obtains the best mAP$^{50\%}$ of 71.8\%, which illustrates the powerful representation ability of attention mechanism. 2) All these method benefits more from larger backbones than the small ones (\eg, ResNet-101 always beats ResNet-50), which is similar to conclusion drawn from semantic segmentation tasks. 3) The one-stage method, SOLOv2, is inferior to these two-stage methods like Mask R-CNN and Cascade Mask R-CNN. This conclusion is also justified in many other datasets. 4) The mAP of each SOTA method are relatively low, \eg, less than 50\%, which manifests the challenge of our \dataset{} dataset.

\begin{table}
\centering
 \caption{Results for fully supervised instance segmentation task.
 The best results are in bold. }
\resizebox{\columnwidth}{!}{
\begin{tabular}{l|l|ll }
\toprule
\cellcolor[HTML]{EFEFEF}\textbf{Methods} & \cellcolor[HTML]{EFEFEF}\textbf{Backbone}  & \cellcolor[HTML]{EFEFEF}\textbf{mAP} & \cellcolor[HTML]{EFEFEF}\textbf{mAP$^{50\%}$} \\
\hline
    \multirow{2}{*}{Mask R-CNN~\cite{He_2017}}              & R50                 & 43.7       & 64.1  \\
                                                            & R101                & 45.1       & 65.3  \\
\cline{1-2}
    \multirow{2}{*}{Cascade Mask R-CNN~\cite{Cai_2019}}     & R50                 & 44.5       & 64.9  \\
                                                            & R101                & 45.3       & 65.3  \\
\cline{1-2}
    \multirow{2}{*}{InstaBoost~\cite{fang2019instaboost}}   & R50                 & 46.1       & 68.0  \\
                                                            & R101                & \textbf{47.8}       & 69.5  \\
\cline{1-2}
     \multirow{2}{*}{SOLOv2~\cite{wang2020solov2}}           & R50                & 28.2       & 45.0  \\
                                                            & R101                & 30.2       & 48.0  \\ 
\cline{1-2}
    \multirow{2}{*}{QueryInst~\cite{Fang_2021_ICCV}}        & R50                 & 45.9       & 69.8  \\
                                                            & R101                & 46.6       & 71.3  \\
\cline{1-2}
    \multirow{2}{*}{Mask2Former~\cite{cheng2021mask2former}}& Swin-T                 & 47.5       & 70.2  \\
                                                            & Swin-S                & \textbf{47.8}       & \textbf{71.8}  \\
\bottomrule
\end{tabular}
}
 \label{tab:instance_fully}
 \\[-0.45cm]
\end{table}
 



\begin{table*}[tb]
\centering
 \caption{Results (mIoU) for semi-supervised semantic segmentation tasks. The best results of each label count are in bold.
 }
 \resizebox{0.67\textwidth}{!}{
\begin{tabular}{c|c|c|llll}
\toprule
\multicolumn{2}{c|}{\cellcolor[HTML]{EFEFEF}\textbf{Methods}}         & \cellcolor[HTML]{EFEFEF}\textbf{Backbone}& \cellcolor[HTML]{EFEFEF}\textbf{1/27}                                  &\cellcolor[HTML]{EFEFEF} \textbf{1/54}           &\cellcolor[HTML]{EFEFEF} \cellcolor[HTML]{EFEFEF}\textbf{1/108}          &\cellcolor[HTML]{EFEFEF} \textbf{1/216}          \\
\hline
\multicolumn{2}{c|}{CutMix~\cite{french2019semi}}             &     & 54.94                                 & 54.05          & 52.58          & 49.73          \\ 
\cline{1-2}
\multicolumn{2}{c|}{ClassMix~\cite{olsson2021classmix}}           &     & 56.10                                 & \textbf{54.79} & 52.92          & 49.81    \\ 
\cline{1-2}
                                  & Stage1 &  & 50.56                                 & 50.72          & 46.72          & 47.71          \\
                                  & Stage2 &  & 55.04                                 & 53.37          & 51.18          & 51.09          \\
\multirow{-3}{*}{Three stage~\cite{ke2020three}}     & Stage3 & \multirow{-5}{*}{R101-DeepLabV2} & {\textbf{56.72}} & 54.31          & \textbf{53.71} & \textbf{53.17} \\
\cline{1-2} \cline{3-3}
\multicolumn{2}{c|}{ST++~\cite{yang2022st++}}                & R50-DeepLabV3+   & 55.34                                 & 50.17          & 46.83          & 48.92          \\

\bottomrule
\end{tabular}
 }
 
  
 \label{tab:semantic_smi}
\end{table*}

\subsection{Semi-Supervised Semantic Segmentation}
\label{sec:setting_semi_seman}


\textbf{Experimental settings. }
%
For the semi-supervised task, we consider different ratios of labelled samples to all the training set: 1/27, 1/54, 1/108, 1/216 where we use  2308 (all of our annotated frames in the training set), 1154, 576 and 288 frames as labelled samples respectively. We then ignore the labels of the remaining labelled samples and use them as unlabelled data. The label-ignored samples together with 59944 unlabelled ones comprise the unlabelled training set. Throughout the experiments, the total number of labelled and unlabelled training samples is fixed to 62252.
%
%
%
The experimental settings of SOTA methods are largely method-dependent. 
We thus use their released code and follow their settings used for Pascal VOC~\cite{Everingham10} dataset. 





\textbf{State-of-the-art methods. }
We implement the following semi-supervised semantic segmentation methods on \dataset{} dataset: CutMix~\cite{french2019semi} and ClassMix~\cite{olsson2021classmix}, Three-stage Self-training~\cite{ke2020three}, ST++~\cite{yang2022st++}. 
%
The former two are the famous semi-supervised semantic segmentation methods proposing augmentation strategies, while the latter two are the latest ones based on pseudo-labels.
CutMix augments an unlabelled image by pasting a randomly copied patch of another sample into it. It is initially proposed for the semi-supervised classification task, but was further extended to segmentation. ClassMix further improves CutMix for segmentation tasks by using the predicted semantic masks for mixing. 
Three-stage Self-training leverages a self-training strategy which utilises pseudo-labels for unlabelled data to train a better model. A three-stage training scheme is then proposed to refine the pseudo-labels. Also following this paradigm, ST++ further exploits the prediction discrepancy of multiple checkpoints to measure the quality of pseudo-labels, with high-quality pseudo-labels used for model training.

\begin{table*}[tb]
\centering
 \caption{Results (mAP) for semi-supervised instance segmentation tasks. The best results of each label count are in bold.
 }
\resizebox{0.75\textwidth}{!}{
\begin{tabular}{c|llllllll}
\toprule
\multirow{2}{*}{\cellcolor[HTML]{EFEFEF}\textbf{Setting}}       & \multicolumn{2}{c}{\cellcolor[HTML]{EFEFEF}\textbf{1/27}}                              & \multicolumn{2}{c}{\cellcolor[HTML]{EFEFEF}\textbf{1/54}}          & \multicolumn{2}{c}{\cellcolor[HTML]{EFEFEF}\textbf{1/108}}          & \multicolumn{2}{c}{\cellcolor[HTML]{EFEFEF}\textbf{1/216}}          \\ 
\cmidrule(lr){2-3} \cmidrule(lr){4-5} \cmidrule(lr){6-7} \cmidrule(lr){8-9}
                       & mAP  & mAP$^{50\%}$ & mAP  & mAP$^{50\%}$ & mAP  & mAP$^{50\%}$ & mAP  & mAP$^{50\%}$ \\ 
\hline
SupOnly                          &   44.3  &   65.0            &   41.4 &   62.0     &   35.0 &   52.7        &   31.7 &   50.4  \\
Semi-sup.~\cite{wang2022noisy}                          &   \textbf{47.4}  &   \textbf{69.1}            &   \textbf{43.9} &   \textbf{66.4}     &   \textbf{37.4} &   \textbf{58.8}        &   \textbf{33.7} &   \textbf{53.9}  \\
\bottomrule
\end{tabular}
}
 \label{tab:instance_smi}
  \\[-0.4cm]
\end{table*}

\textbf{Experimental results. }
Table~\ref{tab:semantic_smi} shows the comparison of these methods on our \dataset{} dataset. We observe that ClassMix is the champion for 1/54 label count while Three Stage wins the other label count settings. The trend for the varied label count is that the mIoU of all these four baselines decreases when the label count reduces. This is reasonable and consistent with what most of the semi-supervised works~\cite{yang2022st++,ke2020three} find, which infers that the boost in performance brought by the increase in unlabelled frames cannot counteract the effect of reducing the labelled frames. 
We also notice that Three Stage and CutMix is less sensitive to varied label count than ST++. For example, when the label count changes from 1/27 to 1/54, the mIoU of CutMix only decreases by 0.89\% while ST++ drops by 5.17\%. 

We further analyze the results for each specific method. Regarding the augmentation-based methods, CutMix is beaten by ClassMix for all the label count settings. This is not surprising as CutMix is not proposed for segmentation tasks while ClassMix improves it for segmentation. ClassMix is also highly competitive with best-performing Three Stage method when there are sufficient labels, like 1/27, 1/54, and 1/108. But it is less effective to handle the label-scarce setting of 1/216. 
%
The first stage training in Three stage~\cite{ke2020three} method is also the supervised only (SupOnly) baseline. Its performance can be improved remarkably (up to 6\% gain) by using the unlabelled data for the second and third stage training, in all the label count ratios. The improvement highlights the importance of the vast amount of unlabelled samples in our \dataset{}. This is also one of the major contributions of our proposed \dataset{}.


Though ST++~\cite{yang2022st++} performs close to the other baselines at the label count of 1/27, its mIoU decreases considerably when the number of labels drops. The significant gap between ST++ and the others may be because the backbone used for ST++ is ResNet-50, whereas all the other baselines employ a deeper backbone,  ResNet-101. Here, ResNet-50 is used for ST++ to save the training cost \textendash  if ResNet-101 is used, ST++ takes significantly longer training time (\eg, 12$\times$ GPU hours) than all the other baselines.

It is also noticeable that the semi-supervised results are significantly worse than the baselines in Table~\ref{tab:semantic_fully} implemented with MMSegmentation. The reason can be two-fold. First, these semi-supervised methods do not use test time augmentation such as flip and multi-scale during inference while MMSegmentation does. Second, the crop size for these methods can be smaller than that in MMSegmentation, \eg, Three Stage\cite{ke2020three} uses a crop size of 321, smaller than 769 in MMSegmentation. 


\subsection{Semi-Supervised Instance Segmentation}
For this task, we use the same label count setting as in~\cref{sec:setting_semi_seman}. Since there are limited works focusing on this task, we only choose the most recent one for evaluation, \ie, the pseudo-label-based method~\cite{wang2022noisy}. This work aims to tackle the noisy boundaries in pseudo-labels by introducing noise-tolerant mask and boundary-preserving map. 
For the experimental setting, we adopt the ResNet-50 based Mask R-CNN as the baseline to perform the training in~\cite{wang2022noisy}. 
We use the the official code released by~\cite{wang2022noisy}, which is also based on MMDetection~\cite{mmdetection}. 

\textbf{Experimental results. }
Table~\ref{tab:instance_smi} shows the comparison of ~\cite{wang2022noisy} with supervised only baselines. The semi-supervised method~\cite{wang2022noisy} consistently surpasses the supervised only (SupOnly) baselines by a clear margin, \eg., $>$2\% mAP and 3\% mAP$^{50\%}$ over SupOnly. The better performance attributes to the large amount of unlabelled data in our \dataset{} dataset. Furthermore, the mAP drops with the decreasing number of labels. The deterioration requires the design of advanced semi-supervised methods to tackle the extreme label-scarce setting on our \dataset{}.

Since ~\cite{wang2022noisy} adopts the same MMDetection~\cite{mmdetection} for the experiments as in fully supervised cases (see Table~\ref{tab:instance_fully}), it is interesting to compare ~\cite{wang2022noisy} with the fully-supervised results under the same framework (and similar settings). For a fair comparison, we only consider the label count of 1/27 (2308 fully labelled training images), where the number of labelled data is the same as the fully-supervised case. We can see that the semi-supervised method~\cite{wang2022noisy} outperforms the Mask R-CNN baseline by 2.4 mAP (47.4 of ~\cite{wang2022noisy} vs. 43.7 in Table~\ref{tab:instance_fully}). Notably, it achieves comparable performance to the transformer-based Mask2Former (47.4 vs. 47.8 mAP). We then observe
that effective exploitation of massive unlabelled data can significantly contribute to  performance gain, given that a large scale unlabelled set are available in our \dataset{}.

\section{Conclusion}
We introduce \dataset{}, the largest traffic flow benchmark for traffic segmentation covering a wide range of cities in India.
The key features of our dataset are two-fold. Firstly, the large variety in classes. Secondly, TrafficCAM is the first dataset to provide a vast amount of unlabelled data. We seek to push forward 
robust and generalisable models that can learn under limited annotations. 
%
We validate our  \dataset{} dataset  on SOTA techniques over four challenging settings: fully-supervised semantic and instance segmentation, and semi-supervised semantic and instance segmentation tasks. 
The experimental results show that even the best performing methods can only achieve unsatisfactory performance on our \dataset{} dataset, with the best mIoU of 66.49\% for semantic segmentation and mAP of 47.8\% for instance segmentation. It hence demonstrates the complexity of 
\dataset{}, opening the door to develop  
more advanced segmentation methods. 


\section*{Acknowledgements}
ZD, AIAR and CBS acknowledge support from the EPSRC grant EP/T003553/1. 
YC and AIAR greatly acknowledge support from a C2D3 Early Career Research Seed Fund and CMIH EP/T017961/1, University of Cambridge. 
LL gratefully acknowledges the financial support from a GSK scholarship and a Girton College Graduate
Research Fellowship at the University of Cambridge. AIAR acknowledges support from CMIH and CCIMI,
University of Cambridge. 
CBS acknowledges support from the Philip Leverhulme Prize, the Royal Society Wolfson Fellowship, the EPSRC advanced career fellowship EP/V029428/1, EPSRC grants EP/S026045/1 and EP/T003553/1, EP/N014588/1, EP/T017961/1, the Wellcome Innovator Awards 215733/Z/19/Z and 221633/Z/20/Z, the European Union Horizon 2020 research and innovation programme under the Marie Skodowska-Curie grant agreement No. 777826 NoMADS, the Cantab Capital Institute for the Mathematics of Information and the Alan Turing Institute. 
The authors also greatly acknowledge KritiKal and Christina Runkel for their insightful discussion.




{\small
\bibliographystyle{ieee_fullname}
\bibliography{egbib}

\begin{thebibliography}{10}\itemsep=-1pt

\bibitem{bae2019spatio}
Bumjoon Bae, Yuandong Liu, Lee~D Han, and Hamparsum Bozdogan.
\newblock Spatio-temporal traffic queue detection for uninterrupted flows.
\newblock {\em Transportation Research Part B: Methodological}, 129:20--34,
  2019.

\bibitem{AAURainSnow}
Chris~H. Bahnsen and Thomas~B. Moeslund.
\newblock Aau rainsnow traffic surveillance dataset, 2018.

\bibitem{Cai_2019}
Zhaowei Cai and Nuno Vasconcelos.
\newblock Cascade r-cnn: High quality object detection and instance
  segmentation.
\newblock {\em IEEE Transactions on Pattern Analysis and Machine Intelligence},
  page 1–1, 2019.

\bibitem{mmdetection}
Kai Chen, Jiaqi Wang, Jiangmiao Pang, Yuhang Cao, Yu Xiong, Xiaoxiao Li,
  Shuyang Sun, Wansen Feng, Ziwei Liu, Jiarui Xu, Zheng Zhang, Dazhi Cheng,
  Chenchen Zhu, Tianheng Cheng, Qijie Zhao, Buyu Li, Xin Lu, Rui Zhu, Yue Wu,
  Jifeng Dai, Jingdong Wang, Jianping Shi, Wanli Ouyang, Chen~Change Loy, and
  Dahua Lin.
\newblock {MMDetection}: Open mmlab detection toolbox and benchmark.
\newblock {\em arXiv preprint arXiv:1906.07155}, 2019.

\bibitem{chen2018encoder}
Liang-Chieh Chen, Yukun Zhu, George Papandreou, Florian Schroff, and Hartwig
  Adam.
\newblock Encoder-decoder with atrous separable convolution for semantic image
  segmentation.
\newblock In {\em Proceedings of the European conference on computer vision
  (ECCV)}, pages 801--818, 2018.

\bibitem{cheng2021mask2former}
Bowen Cheng, Ishan Misra, Alexander~G. Schwing, Alexander Kirillov, and Rohit
  Girdhar.
\newblock Masked-attention mask transformer for universal image segmentation.
\newblock {\em arXiv}, 2021.

\bibitem{mmseg2020}
MMSegmentation Contributors.
\newblock {MMSegmentation}: Openmmlab semantic segmentation toolbox and
  benchmark.
\newblock \url{https://github.com/open-mmlab/mmsegmentation}, 2020.

\bibitem{cordts2016cityscapes}
Marius Cordts, Mohamed Omran, Sebastian Ramos, Timo Rehfeld, Markus Enzweiler,
  Rodrigo Benenson, Uwe Franke, Stefan Roth, and Bernt Schiele.
\newblock The cityscapes dataset for semantic urban scene understanding.
\newblock In {\em Proceedings of the IEEE conference on computer vision and
  pattern recognition}, pages 3213--3223, 2016.

\bibitem{dosovitskiy2020image}
Alexey Dosovitskiy, Lucas Beyer, Alexander Kolesnikov, Dirk Weissenborn,
  Xiaohua Zhai, Thomas Unterthiner, Mostafa Dehghani, Matthias Minderer, Georg
  Heigold, Sylvain Gelly, et~al.
\newblock An image is worth 16x16 words: Transformers for image recognition at
  scale.
\newblock {\em arXiv preprint arXiv:2010.11929}, 2020.

\bibitem{downs1962law}
Anthony Downs.
\newblock The law of peak-hour expressway congestion.
\newblock {\em Traffic Quarterly}, 16(3), 1962.

\bibitem{Everingham10}
M. Everingham, L. Van~Gool, C.~K.~I. Williams, J. Winn, and A. Zisserman.
\newblock The pascal visual object classes (voc) challenge.
\newblock {\em International Journal of Computer Vision}, 88(2):303--338, June
  2010.

\bibitem{fang2019instaboost}
Hao-Shu Fang, Jianhua Sun, Runzhong Wang, Minghao Gou, Yong-Lu Li, and Cewu Lu.
\newblock Instaboost: Boosting instance segmentation via probability map guided
  copy-pasting.
\newblock In {\em Proceedings of the IEEE International Conference on Computer
  Vision}, pages 682--691, 2019.

\bibitem{Fang_2021_ICCV}
Yuxin Fang, Shusheng Yang, Xinggang Wang, Yu Li, Chen Fang, Ying Shan, Bin
  Feng, and Wenyu Liu.
\newblock Instances as queries.
\newblock In {\em Proceedings of the IEEE/CVF International Conference on
  Computer Vision (ICCV)}, pages 6910--6919, October 2021.

\bibitem{french2019semi}
Geoff French, Samuli Laine, Timo Aila, Michal Mackiewicz, and Graham Finlayson.
\newblock Semi-supervised semantic segmentation needs strong, varied
  perturbations.
\newblock {\em arXiv preprint arXiv:1906.01916}, 2019.

\bibitem{gonzalez2008understanding}
Marta~C Gonzalez, Cesar~A Hidalgo, and Albert-Laszlo Barabasi.
\newblock Understanding individual human mobility patterns.
\newblock {\em nature}, 453(7196):779--782, 2008.

\bibitem{He_2017}
Kaiming He, Georgia Gkioxari, Piotr Dollar, and Ross Girshick.
\newblock Mask r-cnn.
\newblock {\em 2017 IEEE International Conference on Computer Vision (ICCV)},
  Oct 2017.

\bibitem{jodoin2014urban}
Jean-Philippe Jodoin, Guillaume-Alexandre Bilodeau, and Nicolas Saunier.
\newblock Urban tracker: Multiple object tracking in urban mixed traffic.
\newblock In {\em IEEE Winter Conference on Applications of Computer Vision},
  pages 885--892. IEEE, 2014.

\bibitem{ke2020three}
Rihuan Ke, Angelica Aviles-Rivero, Saurabh Pandey, Saikumar Reddy, and
  Carola-Bibiane Sch{\"o}nlieb.
\newblock A three-stage self-training framework for semi-supervised semantic
  segmentation.
\newblock {\em arXiv preprint arXiv:2012.00827}, 2020.

\bibitem{li2017vehicle}
Li Li and Xiqun~Michael Chen.
\newblock Vehicle headway modeling and its inferences in
  macroscopic/microscopic traffic flow theory: A survey.
\newblock {\em Transportation Research Part C: Emerging Technologies},
  76:170--188, 2017.

\bibitem{lin2014microsoft}
Tsung-Yi Lin, Michael Maire, Serge Belongie, James Hays, Pietro Perona, Deva
  Ramanan, Piotr Doll{\'a}r, and C~Lawrence Zitnick.
\newblock Microsoft coco: Common objects in context.
\newblock In {\em European conference on computer vision}, pages 740--755.
  Springer, 2014.

\bibitem{liu2021swin}
Ze Liu, Yutong Lin, Yue Cao, Han Hu, Yixuan Wei, Zheng Zhang, Stephen Lin, and
  Baining Guo.
\newblock Swin transformer: Hierarchical vision transformer using shifted
  windows.
\newblock In {\em Proceedings of the IEEE/CVF International Conference on
  Computer Vision}, pages 10012--10022, 2021.

\bibitem{loder2019understanding}
Allister Loder, Lukas Amb{\"u}hl, Monica Menendez, and Kay~W Axhausen.
\newblock Understanding traffic capacity of urban networks.
\newblock {\em Scientific reports}, 9(1):1--10, 2019.

\bibitem{long2015fully}
Jonathan Long, Evan Shelhamer, and Trevor Darrell.
\newblock Fully convolutional networks for semantic segmentation.
\newblock In {\em Proceedings of the IEEE conference on computer vision and
  pattern recognition}, pages 3431--3440, 2015.

\bibitem{loshchilov2017decoupled}
Ilya Loshchilov and Frank Hutter.
\newblock Decoupled weight decay regularization.
\newblock {\em arXiv preprint arXiv:1711.05101}, 2017.

\bibitem{neuhold2017mapillary}
Gerhard Neuhold, Tobias Ollmann, Samuel Rota~Bulo, and Peter Kontschieder.
\newblock The mapillary vistas dataset for semantic understanding of street
  scenes.
\newblock In {\em Proceedings of the IEEE international conference on computer
  vision}, pages 4990--4999, 2017.

\bibitem{olsson2021classmix}
Viktor Olsson, Wilhelm Tranheden, Juliano Pinto, and Lennart Svensson.
\newblock Classmix: Segmentation-based data augmentation for semi-supervised
  learning.
\newblock In {\em Proceedings of the IEEE/CVF Winter Conference on Applications
  of Computer Vision}, pages 1369--1378, 2021.

\bibitem{ramanishka2018toward}
Vasili Ramanishka, Yi-Ting Chen, Teruhisa Misu, and Kate Saenko.
\newblock Toward driving scene understanding: A dataset for learning driver
  behavior and causal reasoning.
\newblock In {\em Proceedings of the IEEE Conference on Computer Vision and
  Pattern Recognition}, pages 7699--7707, 2018.

\bibitem{ren2015faster}
Shaoqing Ren, Kaiming He, Ross Girshick, and Jian Sun.
\newblock Faster r-cnn: Towards real-time object detection with region proposal
  networks.
\newblock {\em Advances in neural information processing systems}, 28, 2015.

\bibitem{strigel2014ko}
Elias Strigel, Daniel Meissner, Florian Seeliger, Benjamin Wilking, and Klaus
  Dietmayer.
\newblock The ko-per intersection laserscanner and video dataset.
\newblock In {\em 17th International IEEE Conference on Intelligent
  Transportation Systems (ITSC)}, pages 1900--1901. IEEE, 2014.

\bibitem{tang2019cityflow}
Zheng Tang, Milind Naphade, Ming-Yu Liu, Xiaodong Yang, Stan Birchfield, Shuo
  Wang, Ratnesh Kumar, David Anastasiu, and Jenq-Neng Hwang.
\newblock Cityflow: A city-scale benchmark for multi-target multi-camera
  vehicle tracking and re-identification.
\newblock In {\em Proceedings of the IEEE/CVF Conference on Computer Vision and
  Pattern Recognition}, pages 8797--8806, 2019.

\bibitem{varma2019idd}
Girish Varma, Anbumani Subramanian, Anoop Namboodiri, Manmohan Chandraker, and
  CV Jawahar.
\newblock Idd: A dataset for exploring problems of autonomous navigation in
  unconstrained environments.
\newblock In {\em 2019 IEEE Winter Conference on Applications of Computer
  Vision (WACV)}, pages 1743--1751. IEEE, 2019.

\bibitem{wang2019deep}
Jindong Wang, Yiqiang Chen, Shuji Hao, Xiaohui Peng, and Lisha Hu.
\newblock Deep learning for sensor-based activity recognition: A survey.
\newblock {\em Pattern recognition letters}, 119:3--11, 2019.

\bibitem{wang2008unsupervised}
Xiaogang Wang, Xiaoxu Ma, and W~Eric~L Grimson.
\newblock Unsupervised activity perception in crowded and complicated scenes
  using hierarchical bayesian models.
\newblock {\em IEEE Transactions on pattern analysis and machine intelligence},
  31(3):539--555, 2008.

\bibitem{wang2020solov2}
Xinlong Wang, Rufeng Zhang, Tao Kong, Lei Li, and Chunhua Shen.
\newblock Solov2: Dynamic and fast instance segmentation.
\newblock {\em Proc. Advances in Neural Information Processing Systems
  (NeurIPS)}, 2020.

\bibitem{wang2022noisy}
Zhenyu Wang, Yali Li, and Shengjin Wang.
\newblock Noisy boundaries: Lemon or lemonade for semi-supervised instance
  segmentation?
\newblock In {\em Proceedings of the IEEE/CVF Conference on Computer Vision and
  Pattern Recognition}, pages 16826--16835, 2022.

\bibitem{xia2018dota}
Gui-Song Xia, Xiang Bai, Jian Ding, Zhen Zhu, Serge Belongie, Jiebo Luo, Mihai
  Datcu, Marcello Pelillo, and Liangpei Zhang.
\newblock Dota: A large-scale dataset for object detection in aerial images.
\newblock In {\em Proceedings of the IEEE conference on computer vision and
  pattern recognition}, pages 3974--3983, 2018.

\bibitem{xie2021segformer}
Enze Xie, Wenhai Wang, Zhiding Yu, Anima Anandkumar, Jose~M Alvarez, and Ping
  Luo.
\newblock Segformer: Simple and efficient design for semantic segmentation with
  transformers.
\newblock {\em Advances in Neural Information Processing Systems},
  34:12077--12090, 2021.

\bibitem{yang2022st++}
Lihe Yang, Wei Zhuo, Lei Qi, Yinghuan Shi, and Yang Gao.
\newblock St++: Make self-training work better for semi-supervised semantic
  segmentation.
\newblock In {\em Proceedings of the IEEE/CVF Conference on Computer Vision and
  Pattern Recognition}, pages 4268--4277, 2022.

\bibitem{zhao2017pyramid}
Hengshuang Zhao, Jianping Shi, Xiaojuan Qi, Xiaogang Wang, and Jiaya Jia.
\newblock Pyramid scene parsing network.
\newblock In {\em Proceedings of the IEEE conference on computer vision and
  pattern recognition}, pages 2881--2890, 2017.

\bibitem{zheng2021rethinking}
Sixiao Zheng, Jiachen Lu, Hengshuang Zhao, Xiatian Zhu, Zekun Luo, Yabiao Wang,
  Yanwei Fu, Jianfeng Feng, Tao Xiang, Philip~HS Torr, et~al.
\newblock Rethinking semantic segmentation from a sequence-to-sequence
  perspective with transformers.
\newblock In {\em Proceedings of the IEEE/CVF conference on computer vision and
  pattern recognition}, pages 6881--6890, 2021.

\end{thebibliography}
}

\end{document}